\documentclass[a4paper]{article}
\usepackage{INTERSPEECH2015,amssymb,amsmath,epsfig,footnote,graphicx,cite}
\usepackage{graphicx}
\usepackage{amssymb,amsmath,bm}
\usepackage{textcomp}
\usepackage[flushleft]{threeparttable}

\ninept

\title{Recurrent Neural Networks with External Memory for \\Language Understanding}

\makeatletter
\def\name#1{\gdef\@name{#1\\}}
\makeatother \name{{\em Baolin Peng$^1$, Kaisheng Yao$^2$
}}

\address{$^1$ The Chinese University of Hong Kong \\
  $^2$Microsoft Research \\
  {\small \tt blpeng@se.cuhk.edu.hk, kaisheny@microsoft.com}
}

\begin{document}

  \maketitle
  \begin{abstract}
Recurrent Neural Networks (RNNs) have become increasingly popular for the task of language understanding. In this task, a semantic tagger is deployed to associate a semantic label to each word in an input sequence. The success of RNN may be attributed to its ability to memorize long-term dependence that relates the current-time semantic label prediction to the observations many time instances away. However, the memory capacity of simple RNNs is limited because of the gradient vanishing and exploding problem. We propose to use an external memory to improve memorization capability of RNNs. We conducted experiments on the ATIS dataset, and observed that the proposed model was able to achieve the state-of-the-art results. We compare our proposed model with alternative models and report analysis results that may provide insights for future research. 
  \end{abstract}

  \section{Introduction}
  \label{sec:introduction}
  
Neural network based methods have recently demonstrated promising results on many natural language processing tasks~\cite{DBLP:journals/jmlr/BengioDVJ03,DBLP:conf/icml/CollobertW08}. Specifically, recurrent neural networks (RNNs) based methods have shown strong performances, for example, in language modeling~\cite{DBLP:conf/interspeech/MikolovKBCK10}, language understanding~\cite{DBLP:conf/interspeech/YaoZHSY13}, and machine translation \cite{DBLP:conf/acl/DevlinZHLSM14,DBLP:conf/emnlp/ChoMGBBSB14} tasks. 

The main task of a language understanding (LU) system is to associate words with semantic meanings \cite{ward1990cmu,DBLP:conf/interspeech/RaymondR07,DBLP:conf/asru/Mori07}. For example, in the sentence "Please book me a ticket from Hong Kong to Seattle", a LU system should tag "Hong Kong" as the departure-city of a trip and "Seattle" as its arrival city. The widely used approaches include conditional random fields (CRFs) \cite{DBLP:conf/interspeech/RaymondR07,DBLP:conf/icml/LaffertyMP01}, support vector machine \cite{DBLP:conf/naacl/KudoM01}, and, more recently, RNNs~\cite{DBLP:conf/interspeech/YaoZHSY13,RNNLUMesnil}.

A RNN consists of an input, a recurrent hidden layer, and an output layer. The input layer reads each word and the output layer produces probabilities of semantic labels. The success of RNNs can be attributed to the fact that RNNs, if successfully trained, can relate the current prediction with input words that are several time steps away. However, RNNs are difficult to train, because of the gradient vanishing and exploding problem~\cite{DBLP:journals/tnn/BengioSF94}. The problem also limits RNNs' memory capacity because error signals may not be able to back-propagated far enough. 

There have been two lines of researches to address this problem. One is to design learning algorithms that can avoid gradient exploding, e.g., using gradient clipping~\cite{DBLP:conf/icml/PascanuMB13}, and/or gradient vanishing, e.g., using second-order optimization methods~\cite{DBLP:series/lncs/MartensS12}. Alternatively, researchers have proposed more advanced model architectures, in contrast to the simple RNN that uses, e.g., Elman architecture~\cite{ElmanRNN}. Specifically, the long short-term memory (LSTM)~\cite{hochreiter1997long, DBLP:conf/icassp/GravesMH13} neural networks have three gates that control flows of error signals. The recently proposed gated recurrent neural networks (GRNN) \cite{DBLP:conf/emnlp/ChoMGBBSB14} may be considered as a simplified LSTM with fewer gates. 

Along this line of research on developing more advanced architectures, this paper focuses on a novel neural network architecture. Inspired by the recent work in \cite{DBLP:journals/corr/GravesWD14}, we extend the simple RNN with Elman architecture to using an external memory. The external memory stores the past hidden layer activities, not only from the current sentence but also from past sentences. To predict outputs, the model uses input observation together with a content retrieved from the external memory. The proposed model performs strongly on a common language understanding dataset and achieves new state-of-the-art results.

This paper is organized as follows. We briefly describe background of this research in Sec. \ref{sec:background}. Section~\ref{sec:model} presents details of the proposed model. Experiments are in section~\ref{sec:experiments}. We relate our research with other works in Sec. \ref{sec:related}. Finally, we have conclusions and discussions in Sec. \ref{sec:conclusions}.

  \section{Background}
  \label{sec:background}
  \subsection{Language understanding}
\label{sec:lu}
A language understanding system predicts an output sequence with tags such as named-entity given an input sequence words. Often, the output and input sequences have been aligned. In these alignments, an input may correspond to a null tag or a single tag. An example is given in Table~\ref{tab:example}.
\begin{table}[htdp]
\begin{center}
\begin{tabular}{|ccccccc|}
\hline
book & a & flight & from & Hong Kong & to & Seattle \\
- & - & - & - & Dpt-city & - & Arv-city \\
\hline
\end{tabular}
\end{center}
\caption{An example of language understanding. Label names have been shortened to fit. Many words are labeled null or '-'.}
\label{tab:example}
\end{table}

Given a $T$-length input word sequence $x_1^T$, a corresponding output tag sequence $y_1^T$, and an alignment $A$, the posterior probability $p(y_1^T|A,x_1^T)$ is approximated by
\begin{equation}
p(y_1^T|x_1^T) \approx \prod_{t=1}^T p(y_t|x_{t-k}^{t+k}),
\end{equation} 
\noindent where $k$ is the size of a context window and $t$ indexes the positions in the alignment. 

\subsection{Simple recurrent neural networks}
\label{sec:srnn}
The above posterior probability can be computed using a RNN. A RNN consists of an input layer $x_t$, a hidden layer $h_t$, and an output layer $y_t$. In Elman architecture~\cite{ElmanRNN}, hidden layer activity $h_t$ is dependent on both the input $x_t$ and also recurrently on the past hidden layer activity $h_{t-1}$. 

Because of the recurrence, the hidden layer activity $h_t$ is dependent on the observation sequence from its beginning. The posterior probability is therefore computed as follows
         \begin{eqnarray}
        p(y_1^T|x_1^T) & \approx  & \prod_{t=1}^T p(y_t|x_1^t) \nonumber \\
        & = & \prod_{t=1}^T p(y_t|h_t,x_t)
        \label{eq1}
        \end{eqnarray}
      \noindent where the output $y_t$ and hidden layer activity $h_t$ are computed as 
      \begin{eqnarray}
      y_t & = & g(h_t), \\
      h_t & = & \sigma(x_t, h_{t-1}). \label{eqn:simplernnht}
      \end{eqnarray}
In the above equation, $g(\cdot)$ is softmax function and $\sigma(\cdot)$ is sigmoid  or tanh function. The above model is denoted as simple RNN, to contrast it with more advanced recurrent neural networks described below. 

\subsection{Recurrent neural networks using gating functions}
\label{sec:gating}
The current hidden layer activity $h_t$ of a simple RNN is related to its past hidden layer activity $h_{t-1}$ via the nonlinear function in Eq.~(\ref{eqn:simplernnht}). The non-linearity can cause errors back-propagated from $h_t$ to explode or to vanish. This phenomenon prevents simple RNN from learning patterns that are spanned with long time dependence~\cite{DBLP:conf/icml/PascanuMB13}. 

To tackle this problem, long short-term memory (LSTM) neural network was proposed in \cite{hochreiter1997long} with an introduction of memory cells, linearly dependent on their past values. LSTM also introduces three gating functions, namely input gate, forget gate and output gate. We follow a variant of LSTM in \cite{DBLP:conf/icassp/GravesMH13}.

More recently, a gated recurrent neural network (GRNN)~\cite{DBLP:conf/emnlp/ChoMGBBSB14} was proposed. Instead of the three gating functions in LSTM, it uses two gates. 

One is a reset gate $r_t$ that relates a candidate activation with the past hidden layer activity $h_{t-1}$; i.e., 
       \begin{eqnarray} 
  \hat{h}_t  = tanh(W_{xh} x_t + W_{hh} ( r_t \odot h_{t-1}))
      \end{eqnarray}
\noindent where $\hat{h}_t$ is the candidate activation. $W_{xh}$ and $W_{hh}$ are the matrices relate the current observation $x_t$ and the past hidden layer activity. $\odot$ is element-wise product. 

The second gate is an update gate $z_t$ that interpolates the candidate activation and the past hidden layer activity to update the current hidden layer activity; i.e., 
\begin{eqnarray}
h_t = (1 - z_t) \odot h_{t-1} + z_t \odot \hat{h}_t. \label{eqn:gatedupdate}
\end{eqnarray}

These gates are usually computed as functions of the current observation $x_t$ and the past hidden layer activity; i.e., 
\begin{eqnarray}
r_t & = & \sigma(W_{xr} x_t + W_{hr} h_{t-1}) \\
z_t & = & \sigma(W_{xz} x_t + W_{hz} h_{t-1}) 
\end{eqnarray} 
\noindent where $W_{xr}$ and $W_{hr}$ are the weights to observation and to the past hidden layer activity for the reset gate. $W_{xz}$ and $W_{hz}$ are similarly defined for the update gate.

  \section{The RNN-EM architecture}
  \label{sec:model}
  
\begin{figure}[t]
\centering
\includegraphics[width=\linewidth]{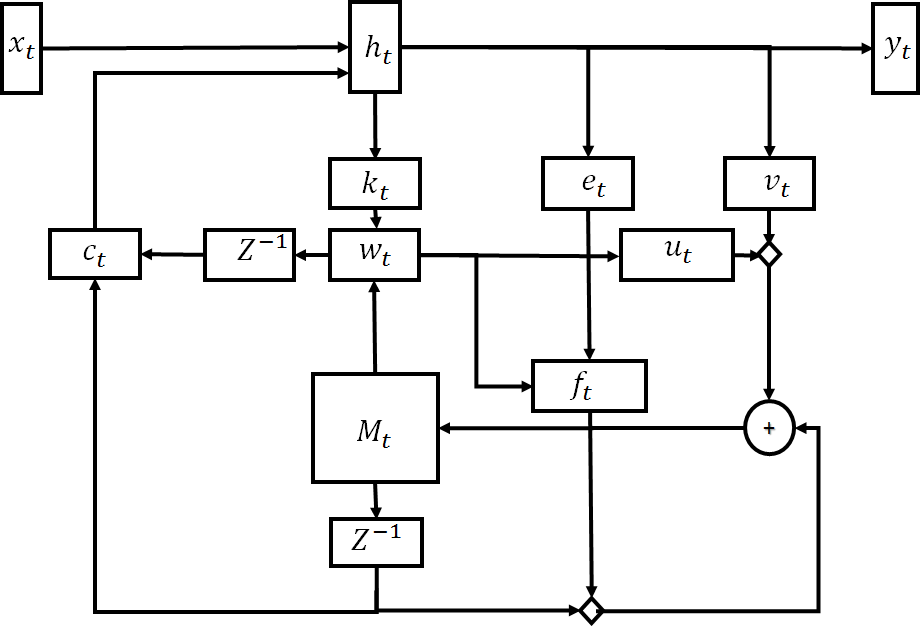}
\caption{The RNN-EM model. The model reads input $x_t$ and outputs $y_t$. Its hidden layer activity $h_t$ depends on the input and the model's memory content retrieved in $c_t$. $f_t$ and $u_t$ are the forget and update gates. $k_t$, $e_t$ and $v_t$ each denote key, erase and new content vector. $M_t$ is the external memory. $w_t$ is the weight and it is a function of $k_t$ and $M_t$. $Z^{-1}$ denotes a time-delay operator. The diamond symbol $\diamond$ denotes diagonal matrix multiplication. \label{fig:ernn}}
\end{figure} 

We extend simple RNN in this section to using external memory. Figure~\ref{fig:ernn} illustrates the proposed model, which we denote it as RNN-EM. Same as with the simple RNN, it consists of an input layer, a hidden layer and an output layer. However, instead of feeding the past hidden layer activity directly to the hidden layer as with the simple RNN, one input to the hidden layer is from a content of an external memory. RNN-EM uses a weight vector to retrieve the content from the external memory to use in the next time instance. The element in the weight vector is proportional to the similarity of the current hidden layer activity with the content in the external memory. Therefore, content that is irrelevant to the current hidden layer activity has small weights. We describe RNN-EM in details in the following sections. 
All of the equations to be described are with their bias terms, which we omit for simplicity of descriptions. We implemented RNN-EM using Theano \cite{Bastien-Theano-2012,bergstra+al:2010-scipy}. 

\subsection{Model input and output}
The input to the model is a dense vector $x_t \in R^{d\times1}$. In the context of language understanding, $x_t$ is a projection of input words, also known as word embedding. 

The hidden layer reads both the input $x_t$ and a content $c_t$ vector from the memory. The hidden layer activity is computed as follows
\begin{equation}
h_t = \sigma(W_{ih}x_t + W_c c_t )
\end{equation}
\noindent where $\sigma(\cdot)$ is tanh function. 
$W_{ih} \in R^{p \times d}$ is the weight to the input vector. 
$c_t \in R^{m \times 1}$ is the content from a read operation to be described in Eq. (\ref{eqn:read}). $W_c \in R^{p \times m}$ is the weight to the content vector. 

The output from this model is fed into the output layer as follows
\begin{equation}
y_t = g(W_{ho} h_t)
\end{equation}
\noindent where $W_{ho}$ is the weight to the hidden layer activity and $g(\cdot)$ is softmax function.

Notice that in case of $c_t = h_{t-1}$, the above model is simple RNN. 

\subsection{External memory read}
\label{sec:read}
RNN-EM has an external memory $M_t \in R^{m \times n}$. It can be considered as a memory with $n$ slots and each slot is a vector with m elements. 
Similar to the external memory in computers, the memory capacity of RNN-EM may be increased if using a large $n$. 

The model generates a key vector $k_t$ to search for content in the external memory. Though there are many possible ways to generate the key vector, we choose a simple linear function that relates hidden layer activity $h_t$ as follows
\begin{equation}
k_t = W_k h_t
\end{equation}
\noindent where $W_k \in R^{m \times p}$ is a linear transformation matrix. 
Our intuition is that the memory should be in the same space of or affine to the hidden layer activity. 

We use cosine distance $K(u,v) = \frac{u \cdot v}{\|u\| \|v\|}$ to compare this key vector with contents in the external memory. The weight for the $c$-th slot $M_t(:,c)$ in memory $M_t$ is computed as follows
\begin{equation}
\hat{w}_t(c) = \frac{\exp \beta_t K(k_t, M_t(:,c))}{\sum_q \exp \beta_t K(k_t, M_t(:,q))}
\end{equation}
\noindent where the above weight is normalized and sums to 1.0. $\beta_t$ is a scalar larger than 0.0. It sharpens the weight vector when $\beta_t$ is larger than 1.0. Conversely, it smooths or dampens the weight vector when $\beta_t$ is between 0.0 and 1.0. We use the following function to obtain $\beta_t$; i.e., 
\begin{equation}
\beta_t = \log(1 + \exp(W_{\beta} h_t))
\end{equation} 
\noindent where $W_{\beta} \in R^{1 \times p}$ maps the hidden layer activity $h_t$ to a scalar. 

Importantly, we also use a scalar coefficient $g_t$ to interpolate the above weight estimate with the past weight as follows:
\begin{equation}
w_t = (1 - g_t) w_{t-1} + g_t \hat{w}_t 
\end{equation}
This function is similar to Eq. (\ref{eqn:gatedupdate}) in the gated RNN, except that we use a scalar $g_t$ to interpolate the weight updates and the gated RNN uses a vector to update its hidden layer activity. 

The memory content is retrieved from the external memory at time $t-1$ using
\begin{equation}
c_t =  M_{t-1} w_{t-1}. \label{eqn:read}
\end{equation}

\subsection{External memory update}
\label{sec:write}
RNN-EM generates a new content vector $v_t$ to be added to its memory; i.e,
\begin{equation}
v_t = W_v h_t
\end{equation}
\noindent where $W_v \in R^{m \times p}$.
We use the above linear function based on the same intuition in Sec. \ref{sec:read} that the new content and the hidden layer activity are in the same space of or affine to each other. 

RNN-EM has a forget gate as follows
\begin{equation}
f_t = 1 - w_t \odot e_t  \label{eqn:forgetgate}
\end{equation}
\noindent where $e_t\in R^{n \times 1}$ is an erase vector, generated as $e_t = \sigma(W_{he} h_t)$. Notice that the $c$-th element in the forget gate is zero only if both read weight $w_t$ and erase vector $e_t$ have their $c$-th element set to one. Therefore, memory cannot be forgotten if it is not to be read. 

RNN-EM has an update gate $u_t$. It simply uses the weight $w_t$ as follows
\begin{equation}
u_t = w_t. \label{eqn:updategate}
\end{equation}
Therefore, memory is only updated if it is to be read. 

With the above described two gates, the memory is updated as follows
\begin{equation}
M_t  = diag(f_t) M_{t-1}  + diag(u_t) v_t \label{eqn:rnnemupdate}
\end{equation}
\noindent where $diag(\cdot)$ transforms a vector to a diagonal matrix with diagonal elements from the vector. 

Notice that when the number of memory slots is small, it may have similar performances as a gated RNN. Specifically, when $n=1$, Eqs. (\ref{eqn:rnnemupdate}) and (\ref{eqn:gatedupdate}) are qualitatively similar.

\section{Experiments}
\label{sec:experiments}
\subsection{Dataset}
In order to compare the proposed model with alternative modeling techniques, we conducted experiments on a well studied language understanding dataset, Air Travel Information System (ATIS)~\cite{dahl1994expanding,wang2006combining,TurSLU}. The training part of this dataset consists of 4978 sentences and 56590 words. There are 893 sentences and 9198 words for test. The number of semantic label is 127, including the common null label. We use lexicon-only features in experiments. 

\subsection{Comparison with the past results}
The input $x_t$ in RNN-EM has a window size of 3, consisting of the current input word and its neighboring two words. We use the AdaDelta method to update gradients~\cite{adadelta}. The maximum number of training iterations was 50.
Hyper parameters for tuning included the hidden layer size $p$, the number of memory slots $n$, and the dimension for each memory slot $m$. The best performing RNN-EM had 100 dimensional hidden layer and 8 memory slots with 40 dimensional memory slot. 
\begin{table}[t]
\begin{center}
\begin{tabular}{|c|c|}
\hline
Method & F1 score \\
\hline 
CRF~\cite{MesnilInterSub13} & 92.94 \\
simple RNN~\cite{DBLP:conf/interspeech/YaoZHSY13} & 94.11 \\
CNN~\cite{DBLP:conf/asru/XuS13} & 94.35 \\
LSTM~\cite{yao2014spoken} & 94.85 \\
GRNN & 94.82 \\
\hline
RNN-EM & 95.25 \\
\hline
\end{tabular}
\end{center}
\vspace{-0.3cm}
\caption{F1 scores (in \%) on ATIS.}
\label{tab:results}
\end{table}

Table \ref{tab:results} lists performance in F1 score of RNN-EM, together with the previous best results of alternative models in the literature. Since there are no previous results from GRNN, we use our own implementation of it for this study. These results are optimal in their respective systems. The previous best result was achieved using LSTM. A change of 0.38\% of F1 score from LSTM result is significant at the 90\% confidence level. Results in Table~\ref{tab:results} show that RNN-EM is significantly better than the previous best result using LSTM. 

\subsection{Analysis on convergence and averaged performances}
\begin{table}[t]
\label{tab:parameters}
\begin{center}
 \begin{threeparttable}
\begin{tabular}{|c|c|c|}
\hline
Model & hidden layer dimension & \# of Parameters \\
\hline
simple RNN & 115 & $\approx 7.4 * 10^3$ \\
LSTM & 50 & $\approx 7.5 * 10^3$ \\
GRNN & 60 & $\approx 7.4 * 10^3$ \\
\hline
RNN-EM\tnote{\textdagger} & 100,40 $\times$ 8 & $\approx 7.3 * 10^3$ \\
\hline
\end{tabular}

\begin{tablenotes}
            \item[\textdagger] 100 dimensional hidden layer, 40 dimensional slot with 8 slots.
        \end{tablenotes}
 \end{threeparttable}
 \end{center}
\vspace{-0.3cm}
\caption{The size of each neural network models.}
\label{tab:parameters}
\end{table}

\begin{figure}[t]
        \centering
        \includegraphics[width=\linewidth]{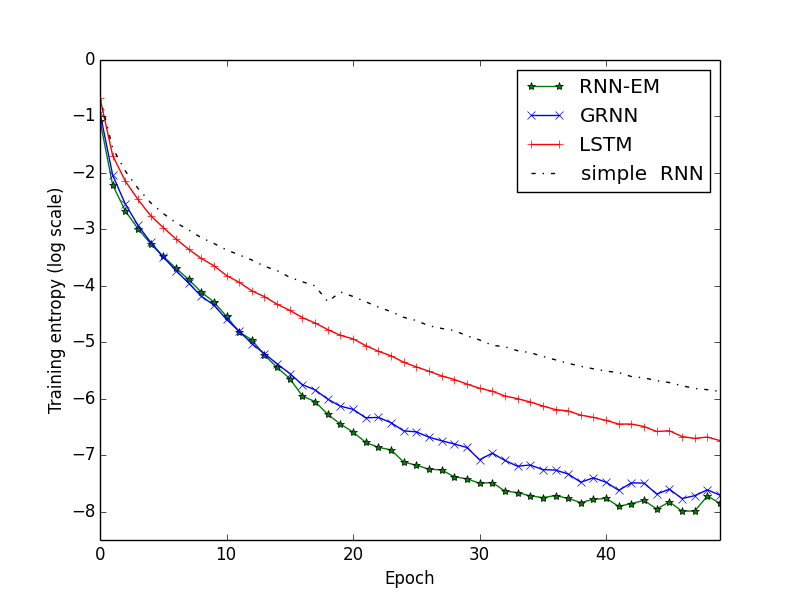}
		\vspace{-0.3cm}
        \caption{{Convergence of training entropy. The entropy value has been converted to its logarithm. }}
        \label{fig:entropy}      
      \end{figure}

Results in the previous sections were obtained with models using different sizes. This section further compares neural network models given that they have approximately the same number of parameters, listed in Table \ref{tab:parameters}. We use AdaDelta~\cite{adadelta} gradient update method for all these models. Figure \ref{fig:entropy} plots their training set entropy with respect to iteration numbers. To better illustrate their convergences, we have converted entropy values to their logarithms. 
 The results show that RNN-EM converges to lower training entropy than other models. RNN-EM also converges faster than the simple RNN and LSTM. 
 
We further repeated ATIS experiments for 10 times with different random seeds for these neural network models. We evaluated their performances after their convergences. Table \ref{tab:statiscs} lists their averaged F1 scores, together with their maximum and minimum F1 scores. A change of 0.12\% is significant at the 90\% confidence level, when comparing against LSTM result. Results in Table~\ref{tab:statiscs} show that RNN-EM, on average, significantly outperforms LSTM. The best performance by RNN-EM is also significantly better than the best performing LSTM. 

\begin{table}[t]
\begin{center}
\begin{tabular}{|c|c|c|c|}
\hline
Method & Max & Min & Averaged \\
\hline
simple RNN & 94.09 & 93.64 & 93.80\\
LSTM & 94.81 & 94.62 & 94.73\\
GRNN & 94.70 & 94.32 & 94.61\\
\hline
RNN-EM & 95.22 &  94.71 &94.96\\
\hline
\end{tabular}
\end{center}
\vspace{-0.3cm}
\caption{The maximum, minimum and averaged F1 scores (in \%) by neural network models.}
\label{tab:statiscs}
\end{table}
 
\subsection{Analysis on memory size}
\label{sec:expmem}
The size of the external memory $M_t$ is proportional to the number of memory slots $n$. We fixed the dimension of memory slots to 40 and varied the number of slots. 
Table~\ref{tab:memsize} lists their test set F1 scores. The best performing RNN-EM was with $n=8$. Notice that RNN-EM with $n=1$ performed better than the simple RNN with 94.09\% F1 score in Table \ref{tab:statiscs}. This can be explained as using gate functions in Eqs. (\ref{eqn:forgetgate}) and (\ref{eqn:updategate}) in RNN-EM, which are absent in simple RNNs. RNN-EM with $n=1$ also performed similarly as the gated RNN with 94.70\% F1 score in Table \ref{tab:statiscs}, partly because of these gate functions. 

Memory capacity may be measured using training set entropy. Table~\ref{tab:memsize} shows that training set entropy is decreased initially with $n$ increased from 1 to 8, showing that the memory capacity of the RNN-EM is improved. However, the entropy is increased with $n$s further increased. This suggests that memory capacity of RNN-EM cannot be increased simply by increasing the number of slots. 
 
\begin{table}[t]
\begin{center}
\begin{tabular}{|c|ccccc|}
\hline
slot number $n$ & 1 & 2 & 4 & 8 & 16\\
\hline 
F1 score & 94.67 & 94.87 & 94.91 & 95.22 & 94.75 \\
entropy$\times 10^3$ & 2.23 & 1.96 & 1.91 & 1.90 & 2.05 \\
\hline
\hline
slot number $n$ & 32 & 64 & 128 & 256 & 512\\
\hline 
F1 score & 94.87 & 94.77 & 94.57 & 94.84 & 94.53\\
entropy$\times 10^3$ & 2.16 & 2.30 & 2.36 & 3.43 & 6.10 \\
\hline

\end{tabular}
\end{center}
\vspace{-0.3cm}
\caption{Test set F1 scores (in \%) and training set entropy by RNN-EM with different slot numbers.}
\label{tab:memsize}
\end{table}      
      

\section{Related works}
\label{sec:related}
The RNN-EM is along the same line of research in ~\cite{DBLP:journals/corr/GravesWD14,DBLP:journals/corr/WestonCB14} that uses external memory to improve memory capacity of neural networks. Perhaps the closest work is the Neural Turing Machine (NTM) work in \cite{DBLP:journals/corr/GravesWD14}, which focuses on those tasks that require simple inference and has proved its effectiveness in copy, repeat and sorting tasks. NTM requires complex models because of these tasks. 
The proposed model is considerably simpler than NTM and can be considered as an extension of simple RNN. Importantly, we have shown through experiments on a common language understanding dataset the promising results from using the external memory architecture.

\section{Conclusions and discussions}
\label{sec:conclusions}
In this paper, we have proposed a novel neural network architecture, RNN-EM, that uses external memory to improve memory capacity of simple recurrent neural networks. On a common language understanding task, RNN-EM achieves new state-of-the-art results and performs significantly better than the previous best result using long short-term memory neural networks. We have conducted experiments to analyze its convergence and memory capacity. These experiments provide insights for future research directions such as  mechanisms of accessing memory contents and methods to increase memory capacity. 

\section{Acknowledgement}
\label{sec:acknowledgement}
The authors would like to thank Shawn Tan and Kai Sheng Tai for useful discussions on NTM structure and implementation.

\bibliographystyle{ieeebib}
\bibliography{mybib}

\end{document}